\pdfoutput=1

\documentclass[11pt]{article}

\usepackage{acl}

\usepackage{times}
\usepackage{latexsym}

\DeclareUnicodeCharacter{05BF}{\relax}

\usepackage{amsmath}
\usepackage{cleveref}
\usepackage{booktabs,makecell}

\usepackage{siunitx, mhchem} 
\usepackage{textcomp}
\usepackage[utf8]{inputenc} 
\usepackage[T1]{fontenc}  
\usepackage{url}  
\usepackage{amsfonts}  
\usepackage{nicefrac}  
\usepackage{microtype} 
\usepackage{multirow,array}
\usepackage{pifont}
\usepackage{graphicx}

\usepackage[T1]{fontenc}

\usepackage{microtype}

\newcommand{\figref}[1]{Fig.~\ref{#1}}
\newcommand{\tabref}[1]{Tab.~\ref{#1}}
\newcommand{\secref}[1]{Sec.~\ref{#1}}

\newcommand{\beas}{\begin{eqnarray*}}
\newcommand{\eeas}{\end{eqnarray*}}
\newcommand{\bea}{\begin{eqnarray}}
\newcommand{\eea}{\end{eqnarray}}
\newcommand{\bes}{\begin{equation*}}
\newcommand{\ees}{\end{equation*}}
\newcommand{\be}{\begin{equation}}
\newcommand{\ee}{\end{equation}}

\newcommand{\tick}{\textcolor{green}{\ding{52}}}
\newcommand{\cross}{\textcolor{red}{\ding{55}}}
\newcommand{\tickcross}{\textcolor{brown}{\ding{70}}}

\makeatletter
\usepackage{xspace}
\usepackage{subcaption}
\def\@onedot{\ifx\@let@token.\else.\null\fi\xspace}
\DeclareRobustCommand\onedot{\futurelet\@let@token\@onedot}

\def\eg{\emph{e.g}\onedot} 
\def\ie{\emph{i.e}\onedot} 
 
 \def\vs{\emph{vs}\onedot}

\title{MeeQA: Natural Questions in Meeting Transcripts}

\author{Reut Apel\thanks{\hspace{0.25cm} Work was done during an internship at Microsoft. Contact: \texttt{\href{mailto:reutapel@campus.technion.ac.il}{reutapel@campus.technion.ac.il}}} \\ 
 Technion -- IIT \\ Microsoft
 \And
 Tom Braude \\ Microsoft\thanks{\hspace{0.25cm}At the time of this work} 
\And  
Amir Kantor \\ Microsoft
ֿ\And
Eyal Kolman \\ Microsoft}

\begin{document}
\maketitle

\begin{abstract}
We present MeeQA, a dataset for natural-language question answering over meeting transcripts. It includes real questions asked during meetings by its participants. The dataset contains 48K question-answer pairs, extracted from 422 meeting transcripts, spanning multiple domains. Questions in transcripts pose a special challenge as they are not always clear, and considerable context may be required in order to provide an answer. Further, many questions asked during meetings are left unanswered. To improve baseline model performance on this type of questions, we also propose a novel loss function, \emph{Flat Hierarchical Loss}, designed to enhance performance over questions with no answer in the text. Our experiments demonstrate the advantage of using our approach over standard QA models.\footnote{Our code and data are available at: \texttt{\href{https://github.com/reutapel/MeeQA}{https://github.com/reutapel/MeeQA}}.}
\end{abstract}

\section{Introduction}
More than a million meetings are held every day in the USA, and employees spend six hours a week attending them on average \citep{mroz2018we, zhong2021qmsum}. Moreover, due to COVID-19 social distancing constraints, remote work has become standard and virtual meetings are now as common as face-to-face \citep{spataro2020future}. Due to this situation, we are bombarded with information, making it more difficult to sort through and distill what is important. In this regard, the rapid improvement of \textit{Automatic Speech Recognitio}n (ASR) can help. Modern ASR incorporates punctuation in the text, such as comma, full stops, and importantly, question marks. The ability to automatically detect questions, coupled with a model capable of extracting the answers to them could help millions of users in alleviating the mental strain and easily distill valuable information from these transcripts. 

\begin{figure}[t]
\includegraphics[width=0.9\linewidth]{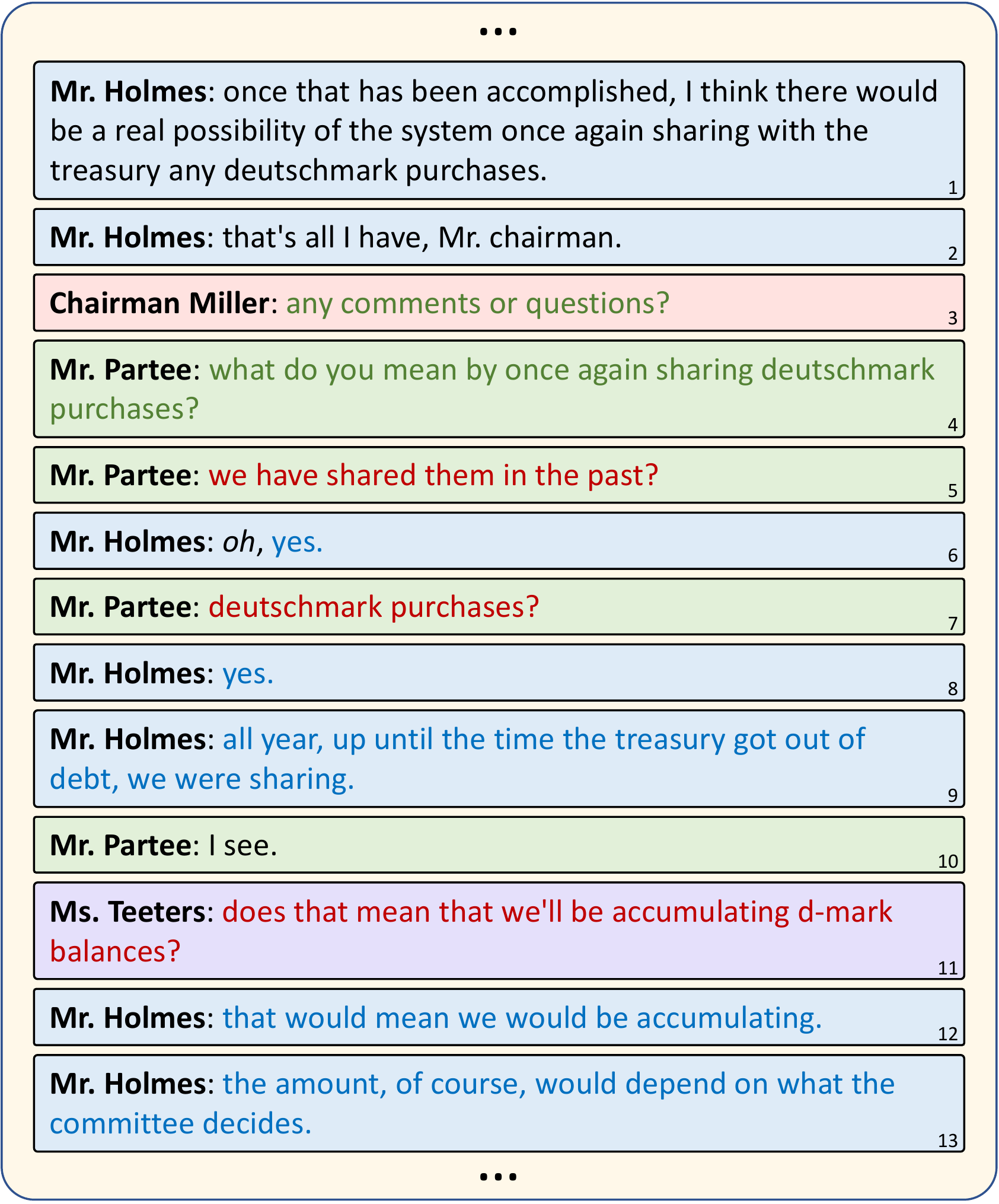}
\centering
\vspace{-0.2cm}
\caption{An example of part of a meeting transcript from MeeQA. Five questions were asked, two ({\color{LimeGreen}green text}) are unanswered, the others ({\color{Maroon}red text}) are answered within the conversation ({\color{NavyBlue}blue text}).}
\vspace{-0.7cm}
\label{fig:dialogue_sample}
\end{figure}

\label{sec:intro}
\begin{table*}[th!]
\centering
\scalebox{0.95}{
\begin{tabular}{ccccccc}
\toprule
\textbf{Dataset} 
& \textbf{\thead{Meeting \\ Transcript}} 
& \textbf{\thead{Spoken \\ Language}} 
& \textbf{\thead{Multi \\ Turn}} 
& \textbf{\thead{Multi \\ Party}}  
& \textbf{\thead{Natural \\ Questions}} 
& \textbf{\thead{Answer \\ Type}} \\ 
\midrule
QAST \citep{qast2009} &  \tickcross &  \tick &  \tick &  \tick &  \cross & E \\
SQuAD 2.0 \citep{squad2} &  \cross &  \cross &  \cross &  \cross &  \cross & E, U \\
QuAC \citep{quac} &  \cross &  \cross&  \tick &  \tick &  \cross & E, U \\
Spoken SQuAD \citep{spokensquad} &  \cross &  \tick &  \cross &  \cross &  \cross & E \\
NQ \citep{naturalquestions} &  \cross &  \cross &  \cross &  \cross &  \tick & E, U, MS \\
CoQA \citep{coqa} &  \cross &  \cross&  \tick &  \tick &  \cross & A*,U\\
TopiOCQA \citep{topiocqa} &  \cross &  \cross &  \tick &  \tick & \tickcross & A, U \\
\midrule
MeeQA (Ours) &  \tick & \tick & \tick &  \tick &  \tick & E, U, MS \\ 
\bottomrule 
\end{tabular}}
\caption{Comparison of MeeQA with other QA datasets. MeeQA contains natural questions and answers from meeting transcript with multiple speakers. Answer types are: Abstractive (A), Abstractive with extractive rationale (A*), Extractive (E), Unanswerable (U), Multi-Span (MS). \tickcross represents that only a proportion of the dataset satisfies the property.}
\vspace{-0.5cm}
\label{tab:datatsets_comparison}
\end{table*}
\textit{Question-answering} (QA) is a task where a machine reads a text passage and then answers questions regarding it. In \textit{conversational question-answer} (CQA), the questions appear in a dialog, and in \textit{spoken question-answer} (SQA), the content, question or both are in spoken form. To advance the capabilities of machines in these tasks, several public datasets have been proposed (see \secref{sec:related_work}). Still, to the best of our knowledge, there are no datasets based on real and natural questions from meeting transcripts.

In this paper, we present a new task, \emph{Meeting Transcripts Question Answering} (MTQA), where a model is required to detect answers to natural questions asked during a meeting, given the subsequent utterances of the conversation. For enabling research on the MTQA task, we present MeeQA, a large-scale dataset for  Meetings Question Answering.
MeeQA contains 48K question-answer pairs from 422 meeting transcripts discussing diverse domains. Each transcript is a set of utterances that were said by multiple speakers. In \figref{fig:dialogue_sample} we show an example of part of a meeting transcript from our dataset with four participants and five questions. 

As our dataset contains text from meeting transcripts with multiple speakers, it has some distinctive characteristics that make it challenging and interesting. First, the text in MeeQA is spoken language. As such, it contains unique language phenomena such as word repetitions, mumblings, use of informal content, as well as informal sentence structure. For instance, in \figref{fig:dialogue_sample} the question in utterance \#5 has an informal structure; utterance \#6 contains the word ``oh'', and utterance \#12 contains word repetition, all of which are typical in spoken language but rare in written text. Further analysis of these phenomena is provided in \secref{subsec:spoken_language}. Second, MeeQA contains meetings from various domains, including board and science groups meetings, each with multiple speakers. Consequently, MeeQA contains diverse questions expressed in a variety of styles. Finally, an essential characteristic of our dataset is that its questions are real and were asked during meetings. As a result, the questions are ``natural'' in the sense that they represent an actual need for information, as opposed to most relevant datasets, where the questions are artificially asked as part of the data collection process.

During real meetings, questions are often left unanswered. Models proposed by previous works to solve QA tasks generally focused on answerable questions and often failed in cases where the answer was not available. Several studies suggested using verification mechanisms during the evaluation process~\citep{bert}; others proposed models that jointly learn the answer and question answerability in multi-task learning setup~\citep{hu2019read+}, and ensemble models were also proposed for this purpose~\citep{RetrospectiveReader}. In this paper, we propose a single model to directly handle unanswered questions, based on a pre-trained language model and a novel loss function, named \emph{flat-hierarchical loss} (FHL). FHL, described in \secref{sec:method}, ties two simple losses, the first for answer span and the other for question answerability prediction. This is accomplished, by leveraging the answerability prediction and ground truth to weight the answer span loss. Based on our results (see \secref{sec:results}), FHL yields models with improved performance on unanswerable questions and comparable performance when the answer is available, which overall provide superior results over the entire dataset. Still, human performance on the unanswerable questions is much better than our best model, highlighting the need for further improvement.
\section{Related Work}
\label{sec:related_work}
\vspace{-0.2cm}
\subsection{Question Answering Datasets}
Question answering is an active research area for which many kinds of tasks have been proposed.
In \emph{reading comprehension} (RC), for example, a model is required to prove its understanding of a given document by answering one or more related questions.
In contrast, in \emph{open-domain question answering} (QA), a model is required to provide answers to questions given one extensive collection of documents (\eg, Wikipedia).
Many different datasets have been curated to advance our understanding of such tasks.

RC datasets such as SQuAD \citep{squad} and NarrativeQA \citep{narrativeqa} consist of questions, relevant documents, and answers which were highlighted in the document. In these datsets, annotators write questions after reading a short text containing the answer.
For the Natural Questions~\citep{naturalquestions} (NQ) and the MS Marco~\cite{msmacro} datasets, the authors collected real queries and responses from Google and Bing respectively with the goal of creating a dataset which is more human-like in nature.

Another kind of task relates to the challenge of question answering in a conversational setup 
commonly named \emph{conversational question answering (CQA)}. In CQA, a model is required to provide answers in a multi-turn QA setup given a context. 
Datasets for this task (\eg CoQA \citep{coqa} and QuAC \citep{quac}) contain 
conversations between a \emph{questioner} and an \emph{answerer} regarding a given external passage. The questioner asks questions about the passage, and the answerer provides the answer.

Recent research focuses on another variant of the question answering family of tasks, namely, \emph{open-domain CQA}, in which the parties are free to discuss various topics during a conversation, drawn from an extensive set of documents. QReCC \citep{QReCC} is a large-scale open-domain CQA and question rewriting dataset, which incorporates an information retrieval subtask. TopiOCQA \citep{topiocqa} is another open-domain CQA dataset, consisting of questions and free-form text answers, with topic switches, based on the Wikipedia corpus. 

Previous researches also applied QA methods to speech transcripts. In \textit{question answering on speech transcripts} (QAST) tasks \citep{qast2007, qast2008, qast2009} passages were taken from transcripts. Then, English native speakers created factual questions about named entities and definitional questions, and the answers were extracted from the transcripts. Spoken SquaAD \citep{spokensquad} is a SQA dataset, in which the documents are the spoken versions of the articles in SQuAD, the questions are original SQuAD questions, and the answer to each question is a span in the document. 

\citet{listening} presented a \textit{machine listening comprehension} (MLC) dataset consisting of recorded spontaneous arguments and formulated questions to confirm or reject these arguments.

In this work, we propose a new task, which we call \emph{Meeting Transcripts Question Answering} (MTQA) (see \secref{sec:intro}), and provide a dataset for it. In this task, the question, the context, and the answers are part of a meeting transcript. Similar to CQA datasets, our setup is multi-party, and similar to QA over speech transcripts, our text is from spoken transcripts. However, our questions and answers are in spoken language and were asked during real meetings by its participants. This combination makes differentiates MeeQA from the aforementioned datasets. 
Furthermore, the questions are asked by different participants and are not only factual, leading to a much more diverse set of questions. To the best of our knowledge, this is the first dataset consisting of natural questions and answers from multi-speakers meetings that were asked in spoken language and converted to text (see \tabref{tab:datatsets_comparison}).
This aspect of our dataset led to much longer questions and answers as well as multi-span answers, which are not common in prior work.

In conclusion, the special characteristics described above differ our dataset from other datasets, thus increasing the challenge posed by previous QA tasks. 

\vspace{-0.2cm}
\subsection{Question Answering Models}
\label{sec:related_work_models}
Most datasets for RC and QA, as well as ours, are characterized by an extractive setup; \ie, detecting the relevant answer within the document.
Previous studies introduced various models to solve these tasks, aiming at extracting single-span answers by jointly learning the start and end tokens of the answer span \citep{DBLP, qanet, bert, xlnet}.

The majority of QA research focuses on answering questions but neglects 
unanswerable ones. \citet{bert} treats questions without answers as having an answer span that starts and ends at the commonly used \texttt{[CLS]} token. \citet{RetrospectiveReader} proposed a retrospective reader that consists of a sketchy reader predicting question \emph{answerability}, followed by an intensive reader that extracts answers and combines them with the sketchy prediction to yield a final answer. In contrast, we propose a method to directly handle unanswerable questions through a unique loss function described in \secref{subsec:fhl}.

\vspace{-0.2cm}
\section{MeeQA Dataset}
\label{sec:dataset}
Our dataset is focused on natural questions and answers that were extracted from professional meetings. Each instance in our dataset contains a question asked during a meeting, three utterances before the question, and 60 utterances after it. The answers are spans within the utterances that followed the question. Data collection was performed in two stages described below: Aggregating meeting transcripts from multiple sources and question extraction and annotation.
\subsection{Data Collection}
MeeQA is composed of different meeting types, such as product design, financial, and board meetings. To achieve this, we aggregated meeting transcripts from multiple sources:

\noindent\textbf{Augmented Multi-party Interaction (AMI)
meeting corpus}~\citep{AMI} - a dataset of 100 hours from 159 recorded meetings about product design in an industrial setting. Some of the meetings are naturally occurring, and some are elicited in which a design team, taking a design project from kick-off to completion over a day. It contains the manually annotated meeting transcripts and their meeting summaries. 

\noindent\textbf{ICSI meeting corpus}~\citep{ICSI} - an academic meeting dataset that consists of 75 weekly group meetings at the International Computer Science Institute (ICSI) in Berkeley.

\noindent\textbf{Board Meetings} - another interesting domain of meetings which contains formal discussions about a wide range of topics and are publicly available. We include board meetings transcripts of the Legal Services Corporation (LSC)\footnote{Legal Services Corporation: America’s Partner for Equal Justice} as well as other transcribed board meetings that were extracted from available public resources.\footnote{We use meetings from the California Citizens Redistricting Commission (CCRC) corpus (\url{https://wedrawthelines.ca.gov/transcripts/}), Federal Open Market Committee (FOMC) corpus (\url{https://www.federalreserve.gov/monetarypolicy/fomc_historical_year.htm}), and Professional Learning Community (PLC) corpus.}

\vspace{-0.1cm}
\subsection{Annotation Pipeline}
\label{annotation_pipeline}
Questions are common in meeting transcripts. Specifically, about 10\% of all the utterances considered for MeeQA include at least one sentence ending in a question mark, and only such sentences were considered question candidates. 
The collected corpus was divided into tasks, each task includes 20 question candidates from the same meeting. Each candidate was initially presented to the judges with ten utterances before and ten utterances after it. If needed, judges were also given the option to expand the number of displayed utterances. Examples of the annotation interface are provided in the Appendix. Each task was annotated by three different judges following a three steps methodology: 

\noindent\textbf{Question Verification} - judges are asked if the question candidate requires an answer. If the judges respond positively, they proceed to the next step, otherwise, they proceed to the next candidate. 

\noindent\textbf{Question Type Labeling} - The goal here is to enrich our dataset with information about the question. In this step, the judge is asked four questions regarding the question itself. We ask the judges about the question comprehension, whether it is self-contained, is it a yes/no question, and whether it is relevant and informative after the meeting.

\noindent\textbf{Answer Marking} - judges are asked to mark the answer in the transcript.  Importantly, though we did not limit the judges to highlighting only within the utterances after the question, we did exclude answers located in the utterances before the question from our final dataset. The judges were also not limited to highlighting complete words; therefore, we completed partially highlighted words as part of our cleaning process.

\begin{figure}[t]
\includegraphics[width=1\linewidth]{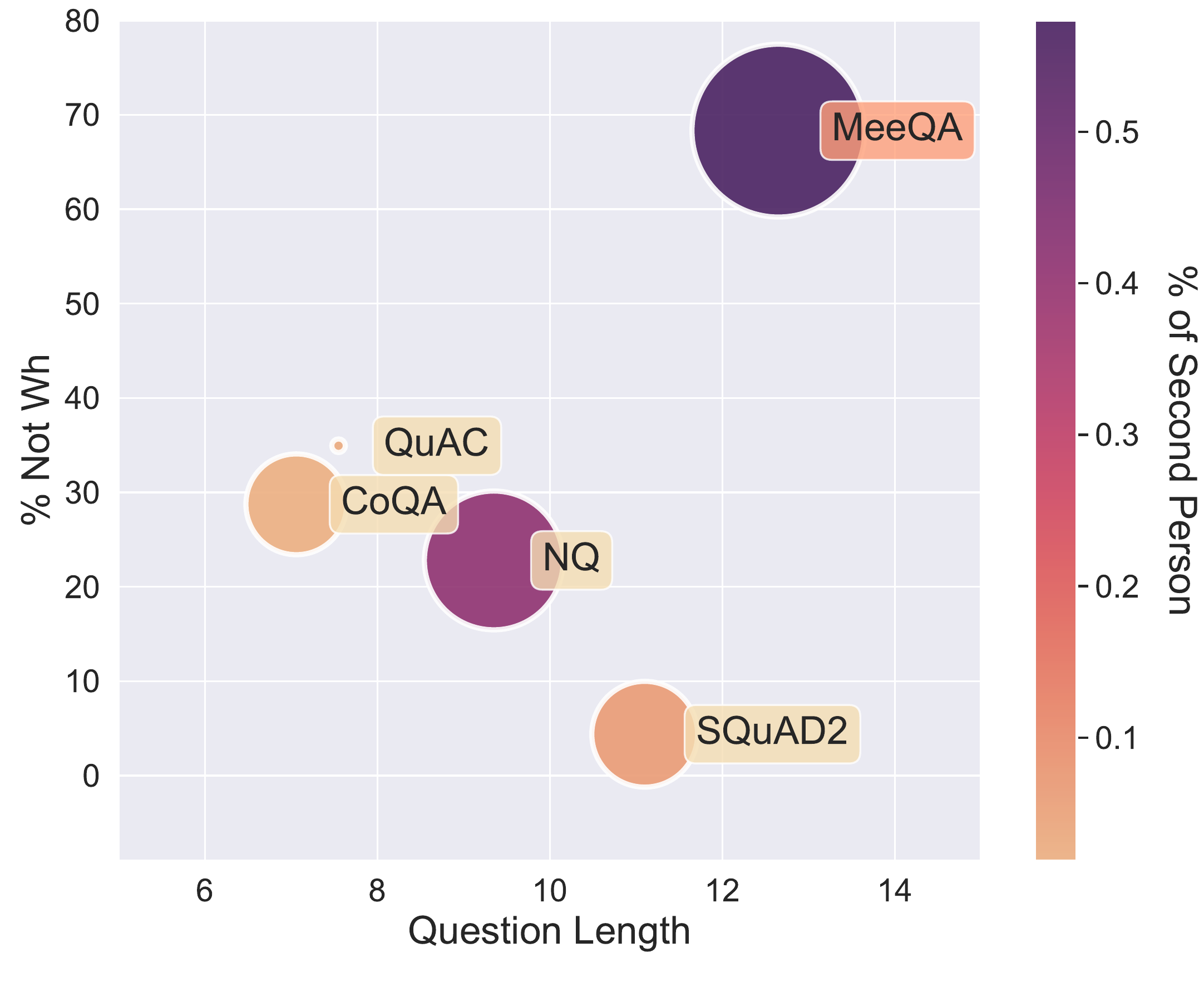}
\vspace{-0.8cm}
\caption{Datasets comparison in four dimensions: the average question length (vertical axis), the percentage of non-Wh-questions (horizontal axis), the percentage of present tense questions out of all present and past tense questions (size), and the percentage of second-person questions out of all the second and third-person questions (color).}
\vspace{-0.5cm}
\label{fig:dataset_comparison}
\end{figure}

\section{Dataset Statistics and Analysis}
\label{sec:dataset_stat}
\vspace{-0.2cm}
We present an analysis MeeQA in \tabref{tab:data_statistics}, composed of 19,142 unique questions, labeled by ~3 judges each, from 422 different meetings discussing various topics. Thus, the total number of question-answer pairs is 48,009. The dataset is partitioned randomly into a training, validation and test set, by an approximate 70/15/15 split (approximated as we constrain questions from the same meeting to belong to the same split to avoid data leakage). In this section we describe MeeQA properties, and demonstrate the unique challenge it introduces for QA tasks.

\begin{table}[t]
\scalebox{0.82}{
\centering
\begin{tabular}{llll|l}
\toprule
 & \textbf{Train} & \textbf{Dev} & \textbf{Test} & \textbf{Overall} \\
 \midrule
\# Meetings & 293 & 61   & 68 & 422   \\
\# Unique Questions & 13,426 & 2,892  & 2,824 & 19,142   \\
\# Q/A Pairs  & 33,690 & 7,188  & 7,131 & 48,009   \\
\% Unanswerable  & 23.0   & 25.4 & 24.7  & 23.6  \\ 
\bottomrule
\end{tabular}}
\caption{Data statistics. Number of meetings, number of unique candidate questions, and number of questions that were tagged with and without answers.}
\vspace{-0.5cm}
\label{tab:data_statistics}
\end{table}

\subsection{Spoken Language}
\label{subsec:spoken_language}
As MeeQA is based on spoken language, it presents unique characteristics not found on most datasets. We conduct an analysis of the prevalence of different phenomenon based on sampling two question-answer pairs from each meeting in the development set, followed by manual annotation. \emph{Informal Content} (\eg, use of ``yeah'' or ``gonna'') is the most frequent phenomenon which constitutes 29.2\% of the question-answer pairs. Other aspects of spoken language include \emph{Filler Words} (\eg, ``hmm'' or ``huh''), found in 25\% of the sampled sentences and \emph{Word Repetition} (\eg, ``what is what is the cost?'') present in 14.2\%.
Moreover, some sentences are not formal due to the lack of punctuation or incorrect sentence structure. Our analysis shows that 32.5\% of the question-answer pairs contain at least one informal sentence, classified as \emph{Informal Structure}. In total, 59.2\% of the pairs include at least one of the aforementioned phenomena, where each example can be annotated with more than one phenomenon.

\subsection{Question Properties}
The utterances that are auto-transcribed during meetings are not always comprehensible. Specifically, 7\% of the questions in MeeQA were labeled as non-comprehensible by our judges. These questions might cause difficulties for answer detection for both humans and machines. Interestingly, 37\% of the questions are not self-contained, \ie, they do not contain all the required information to understand the question and answer it. This characteristic is also common in CQA datasets such as CoQA~\cite{coqa}, as there are dependencies between the question and the previous utterances. Further, 56\% of the questions were tagged as relevant also after the meeting, and 48\% of them were tagged as yes/no questions.  

Apart from the differences between MeeQA and other datasets mentioned in \secref{sec:related_work}, the questions on MeeQA have several unique properties. Specifically, we focus on four characteristics and compare MeeQA to four large-scale QA and RC datasets.\footnote{publicly available at \url{https://huggingface.co/datasets}} \figref{fig:dataset_comparison} illustrates this comparison in four dimensions. The vertical axis represents the average question length, and the horizontal axis represents the percentage of non-\emph{Wh}-questions. While other datasets contain more than half \emph{Wh}-questions with an average length of at most 11 tokens, almost 70\% of the questions in our datasets are not \emph{Wh}-questions with an average length of more than 12 tokens. The other two characteristics are represented by the size and color of each data point. Specifically, the size represents the percentage of present tense questions out of all present and past tense questions. The color represents the percentage of second-person questions out of all the second and third-person questions. To calculate these two measures, we use the Stanford NLP POS tagger \citep{stanfordnlp} to tag each token with its person and tense, and finally tag each question based on the majority of the person and tense over its words. \figref{fig:dataset_comparison} shows that compared to other datasets, MeeQA contains significantly more present tense and second person questions, which fits the nature of natural and non-factual questions. For example, ``do you'' is a common first bigram in MeeQA questions.

\subsection{Answer Properties}
We analyze answers in the development set which contains 6,416 questions with answers, comprising over 70\% of the development set.
Transcripts are partitioned to utterances, answers in MeeQA span 2.13 such utterances on average, where 60\% of them are contained in a single one.
Furthermore, 69\% of the answers include, but are not limited to, the first utterance after the question, which is expected in the context of meetings and differentiates MeeQA from other QA and RC datasets.
Nevertheless, the agreement between judges is partial. We calculate Krippendorff's reliability measure, $\alpha$ \cite{hayes2007answering} to formally measure judge agreement. When considering annotations at the level of individual words in each transcript, the overall score for all examples and all annotators is $\alpha = 0.555$. Low agreement score is expected as transcripts are inherently noisy thus judgement is challenging even for humans.

Additionally, answers on MeeQA are significantly longer with an average of 31.1 tokens compared to 3.2 in SQuAD~\cite{squad}, and 2.7 in CoQA~\cite{coqa}. These differences are not surprising as most of the answers in these datasets are either entities, numbers, or Yes/No answers (for CoQA). Interestingly, our average answer length is over twice that of QuAC~\cite{quac} (31.1 \vs 15). Compared to MeeQA, which derives answers from natural speech of meeting participants, answers in QuAC are derived from specific passages, and therefore can be less detailed.
\section{Method}
\label{sec:method}
As discussed in \secref{sec:related_work_models}, many models that solve QA tasks focus on answerable questions and might falter in their predictions for the unanswerable ones. As questions asked during meetings are many times left unanswered, a desired capability of a model in our setup is detecting when an answer is not available. To that end, we propose a novel loss function that combines two tasks, span prediction and detecting if an answer is available. In the following section we formally introduce the \emph{MTQA} task and define our \emph{flat-hierarchical loss}.
\begin{table*}[ht]
\centering
\scalebox{0.85}{
\begin{tabular}{c|cccccc|cccccc}
\toprule
& \multicolumn{6}{c|}{\textbf{Dev}}& \multicolumn{6}{c}{\textbf{Test}}\\ \midrule
\multirow{2}{*}{\textbf{Model}} & \multicolumn{2}{c}{\textbf{All Data}} & \multicolumn{2}{c}{\textbf{HasAns}} & \multicolumn{2}{c|}{\textbf{NoAns}} & \multicolumn{2}{c}{\textbf{All Data}} & \multicolumn{2}{c}{\textbf{HasAns}} & \multicolumn{2}{c}{\textbf{NoAns}} \\ \cmidrule{2-13} 
& \textbf{EM} & \textbf{F1} & \textbf{EM}& \textbf{F1}& \textbf{EM}& \textbf{F1}& \textbf{EM} & \textbf{F1} & \textbf{EM}& \textbf{F1}& \textbf{EM}& \textbf{F1}\\ \midrule
\textbf{Human Performance} & 53.0 & 67.3 & 46.4 & 62.8 & 88.3 & 88.9 & 53.9 & 68.3 & 47.8 & 64.3 & 88.2 & 88.6 \\ \midrule
\textbf{First Utterance} & 40.2 & 54.7 & 46.2 & 60.4 & 3.1 & 19.7 & 41.1 & 55.7 & 47.3 & 61.2 & 2.4 & 21.2 \\ \midrule
\textbf{BERT} & 61.9& 75.2& 59.9 & \textbf{74.8} & 73.9 & 77.7 & 62.0& 74.7& 60.6 & \textbf{74.5} & 70.7 & 76.1\\
\textbf{BERT + FHL} & \textbf{62.9}& \textbf{75.7}& \textbf{60.2} & 74.6 & \textbf{79.6} & \textbf{82.5} & \textbf{62.3}& 74.7& \textbf{60.8} & 74.4 & \textbf{72.2} & \textbf{76.9}\\ \midrule
\textbf{DeBERTa} & 51.8& 75.5& 48.5 & \textbf{75.1} & 72.1 & 77.9 & 53.8& \textbf{75.9}& 51.2 & \textbf{75.9} & 69.8 & 76.0\\
\textbf{DeBERTa + FHL} & \textbf{53.9}& \textbf{76.0}& \textbf{50.0} & 74.8 & \textbf{77.9} & \textbf{83.2} & \textbf{55.2}& 75.2& \textbf{52.0} & 74.3 & \textbf{75.5} & \textbf{80.9}\\ \midrule
\textbf{ALBERT}& 63.1& 75.9& \textbf{60.5} & \textbf{75.0} & 79.2 & 82.0 & \textbf{63.6}& \textbf{75.9}& \textbf{61.4} & \textbf{75.0} & 77.5 & 81.5\\
\textbf{ALBERT + FHL}& \textbf{64.0}& \textbf{76.4}& 60.3 & 74.5 & \textbf{86.5} & \textbf{88.0} & 63.2& 75.5& 60.4 & 74.1 & \textbf{80.7} & \textbf{84.4}\\ 
\bottomrule
\end{tabular}}
\caption{Baseline comparison results. Models (standard span prediction objective \vs FHL), baselines and human performance on the development and the test data. We consider a question to be unanswerable if at least half of the judges tagged it as such.}
\vspace{-0.5cm}
\label{tab:main_results}
\end{table*}

\subsection{Task Definition} \label{subsec:task_definition}
Let $U=\{u_m | 1\leq m \leq M\}$ be a set of $M$ ordered utterances $u_m$ which belong to a single meeting. Note that an utterance can contain multiple sentences. Let $Q$ be a sentence in utterance $u_q$, which ends with a question mark, hereby $q$ is the index of the utterance to which question $Q$ belongs. Additionally, let $u_{q_{-}}$ denote the prefix of $u_q$ ending at the start of question $Q$ and $u_{q_{+}}$ the suffix of $u_q$ after removing $u_{q{-}}$ and $Q$. Thus, $u_q=u_{q_{-}}Q u_{q_{+}}$, \ie, the concatenation of the aforementioned strings.

Given a question $Q$, $k$ utterances before the question $U_B=(u_{q-k},\dots,u_{q-1},u_{q_{-}})$, and $l$ utterances after it $U_A=(u_{q_{+}},u_{q+1},\dots,u_{q+l})$, two goals are required. First, predict whether an answer exists for $Q$, hereby $y_{HA}\in \{0,1\}$ denotes the label for answerability (\ie, \textbf{H}as \textbf{A}nswer). Second, find two indices $y_S$ and $y_E$ corresponding to \textbf{S}tart and \textbf{E}nd tokens of a span in the concatenation of $U_A$ and $y_S \leq y_E$.


\begin{figure}[t]
\centering
\scalebox{0.7}{
\includegraphics[width=1\linewidth]{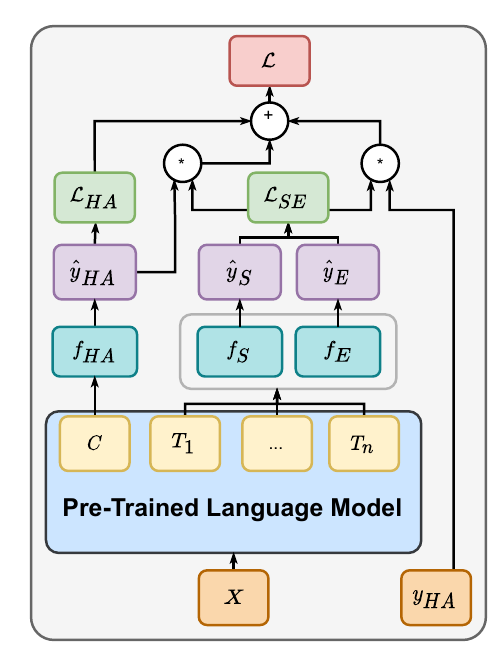}}
\caption{Computation of Flat-Hierarchical loss.}
\vspace{-0.5cm}
\label{fig:fhl}
\end{figure}
\label{subsec:model}
\paragraph{Base Model} 
Following the practice introduced by BERT~\citep{bert}, we concatenate $Q$, $U_A$, and $U_B$ into a single input sequence $X$. We first define two strings, $S_B$  -  text up-to and including $Q$, \ie,  $u_{q-k}\dots u_{q-1}u_{q_{-}}Q$; and $S_A$ - text proceeding $Q$, \ie,  $u_{q_{+}}u_{q+1}\dots u_{q+l}$. The final input $X$ is obtained as follows:
$\texttt{[CLS]} S_B \texttt{[SEP]} S_A \texttt{[SEP]}$.
In essence, we use the standard \texttt{[CLS]} and \texttt{[SEP]} tokens, and split the meeting utterances up-to and after the question $Q$. We append a special token ``\&'' denoting the start of the utterance, followed by a speaker token to each utterance $u_t$ and then tokenize the resulting string. To normalize across the different data sources, we convert the speaker roles and names to ``SPEAKER\_$Z$,'' where $Z$ is an index of the speaker, and the questioner is always SPEAKER\_$0$.

We first embed the sequence $X$ via a pre-trained transformer encoder (\eg BERT). The resulting representation is then  fed through three independent linear classifiers. The first, $f_S$, outputs the probability of each word in the sequence to be the start of the answer span. Similarly, $f_E$ predicts the end of the answer span. Finally, $f_{HA}$ uses the first special token, \texttt{[CLS]}, to predict the question \emph{answerability}, \ie, is the question answerable given the input $X$. Formally, we denote the transformer encoder output for each token in the sequence by $T_1$, ..., $T_n\in \mathbb{R}^{d}$, and for the token \texttt{[CLS]} by $C\in \mathbb{R}^{d}$. Given $T=(T_1, ..., T_n)$ and $C$ we get:
\vspace{-0.1cm}
\begin{align}
\hat{y}_S \propto \text{SoftMax}(W_S\cdot T) \label{eq:softmax1}\\ 
\hat{y}_E \propto \text{SoftMax}(W_E\cdot T) \label{eq:softmax2}\\
\hat{y}_{HA} \propto \text{SoftMax}(W_{HA}\cdot C) \label{eq:softmax3}
\end{align}

where $W_S$, $W_E$, $W_{HA}\in \mathbb{R}^{d}$ are trainable weights, $\hat{y}_S$ and $\hat{y}_E$ are the model prediction for the start and end tokens of the answer span, and $\hat{y}_{HA}$ is the model prediction for the question answerability. The dimension of all the $\hat{y}$'s is two.

\paragraph{Flat-Hierarchical Loss Function}\label{subsec:fhl}
When people are asked to highlight an answer to a question in a given passage, they first read the question and passage and decide if an answer is found in the text \citep{Literacy, human}. Only if they deem that it is, they proceed to marking the relevant span. Mimicking this hierarchical logic, we propose a novel loss function named \emph{flat-hierarchical loss} (FHL), illustrated in \figref{fig:fhl}. To calculate FHL we first compute two standard losses. The first, $\mathcal{L}_{SE}$, for start-end prediction computed as the average of two cross-entropy losses of the start and end span classifiers. The second, $\mathcal{L}_{HA}$, for Has-Answer prediction, by the cross-entropy loss for the answerability classifier. Once the two terms are computed, we compute the final loss, $\mathcal{L}$, as follows:

\vspace{-0.5cm}
\begin{align}\label{eq:hierarchical_loss}
    \begin{split}
        \mathcal{L} = \frac{1}{N} \sum\limits_{i=1}^{N} \biggr( \alpha \cdot\mathcal{L}_{HA}(y_{HA}, \hat{y}_{HA}) + \\
        \beta \cdot \hat{y}_{HA}\cdot \mathcal{L}_{SE}(y_S,\hat{y}_{S}, y_E,\hat{y}_{E}) + \\
        \gamma \cdot y_{HA}\cdot \mathcal{L}_{SE}(y_S,\hat{y}_{S}, y_E,\hat{y}_{E}) 
        \biggr)
    \end{split}
\end{align}

where $\alpha$, $\beta$, and $\gamma$ are weights that calibrate the importance of each component in the function, and are selected to maximize F1 over the development set. $N$ is the number of examples,  
$\hat{y}_{HA}$ and $y_{HA}$ are the model answerability prediction, and ground truth labels respectively, and $\hat{y}_{S}$, $y_{S}$, $\hat{y}_{E}$ and $y_{E}$ are model predictions and labels for start and end indices.

Intuitively, if the model calculates a low probability of answerability and if the question is indeed  unanswerable, the weights of $\mathcal{L}_{SE}$ would be diminished. That is, if there is no answer, the prediction of the Start and End heads is less important. Using FHL, the model trains to predict the span answer conditioned on probability it assigns for the existence of an answer. Moreover, this formulation disentangles the role of the \texttt{[CLS]} token from span prediction.
\paragraph{Threshold-based Answer Decision}
Our model provides two independent predictions: whether the question has an answer and answer span candidates. To produce the final prediction, we propose a threshold-based answer decision method that combines these two predictions, inspired by previous studies \citep{bert, RetrospectiveReader}. In particular, we define the score of each candidate span from index $i$ to index $j$ as follows: $score_{i,j} = W_S \cdot T_i +W_E \cdot T_j$. We then compute the probability of each candidate that is no longer than $m$ tokens being the answer span by a softmax over all the candidates, and $P_{best} = \max_{j \geq i} \frac{e^{score_{i,j}}}{\sum{e^{score_{m,n}}}}$ is the probability of the final answer span. Finally, we predict that there is no answer when $\hat{y}_{HA} \leq \tau_1 \wedge P_{best} \leq \tau_2$, where the thresholds $\tau_1$ and $\tau_2$, and the number of tokens $m$ are selected to maximize F1 over the development set.

\section{Experiments}\label{sec:results}
\vspace{-0.2cm}
\subsection{Implementation Details}\label{sec:imp_details}
\noindent\textbf{Data Representation}
For all experiments in this section we merge consecutive utterances of the same speaker to a single utterance. For each question we consider 1 utterance before and 60 utterances after the question (\ie, $k=1$ and $l=60$ in \secref{subsec:task_definition}). The maximum input length in all experiments is 512 tokens. We truncate long sequences and pad shorter ones to the maximum input length. An analysis of various parameters and merging strategies is provided in the Appendix.
\begin{table}[t]
\centering
\begin{tabular}{c|cc}
\toprule
\textbf{Model} & \textbf{EM} & \textbf{F1}\\ 
\midrule
\textbf{No-HA}& 47.0& 54.1\\
\textbf{No-PSE}& 62.7& 74.9\\
\textbf{No-LSE}& 63.4& 75.2\\ \hline
\textbf{Full model}& 63.2 & 75.5 \\
\bottomrule
\end{tabular}
\vspace{-0.3cm}
\caption{Ablation model results for the test data with ALBERT}
\vspace{-0.7cm}
\label{tab:ablation_results}
\end{table}

\noindent\textbf{Training Setup}
We use the HuggingFace Transformers library \citep{transformers_huggingface}
to implement our models. We consider three pre-trained models: BERT~\cite{bert}, DeBERTa~\cite{deberta}, and ALBERT~\cite{albert}, to which we stack three linear classifiers (see Eq. \eqref{eq:softmax1}, \eqref{eq:softmax2} and \eqref{eq:softmax3}), and implement the loss function as described in Eq. \eqref{eq:hierarchical_loss}.
All models are trained for two epochs and optimized via the ADAMW optimizer \citep{adam}. We employ a batch size of 8 and a learning rate of $3 \cdot 10^{-5}$.  
After training, we tune $\alpha$, $\beta$ and $\gamma$ parameters (See \secref{subsec:fhl}) considering the values $\{0.7, 0.8\}$, $\{0.2, 0.3\}$, and $\{0.7, 0.8\}$ respectively. For the answer decision step, we consider the values $\{0.6, 0.7\}$ and $\{0.8, 0.9\}$ for $\tau_1$ and $\tau_2$ respectively. For all models, we consider the values $\{200, 250\}$ for the maximum answer length $m$. All tuning steps described above are performed on our development set and then applied to the test set during evaluation.

\vspace{-0.3cm}
\subsection{Evaluation Metrics}
\vspace{-0.2cm}
Following SQuAD \citep{squad}, we evaluate our performance by two metrics; Exact Match (EM) and macro-average F1 score. The answers in our dataset are characterized by long sentences (as opposed to short evidence); thus, we opt for calculating F1-score at the level of individual (ordered) words, instead of considering a bag-of-words. Recall and precision are computed by considering the overlap between predicted word indices and ground truth answers. Questions without an answer are assigned an F1 score of 1 if the model predicted ``no answer'' and 0 otherwise. In SQuAD, each model prediction is compared to $n$ annotated answers for computing these metrics, and the maximum score is chosen as the question score. However, for computing human performance, each human annotation is compared to the other $n-1$ annotations. Thus, the human performance and the model results are not comparable. To moderate this bias, we compare the model results to $n$ subsets of $n-1$ annotated answers, and the final scores are calculated as the average of the maximum scores of all subsets, similar to \citet{coqa} and \citet{quac}.

\vspace{-0.2cm}
\subsection{FHL Evaluation}
\noindent\textbf{Baseline Comparison Analysis}
We conduct a comparative evaluation to assess the significance of our novel FHL loss (See \secref{subsec:fhl}). For each of the three baseline models (See \secref{sec:imp_details}) we compare the performance when optimizing for the standard span prediction objective against the performance when optimizing using FHL. Results of these experiments are reported in \tabref{tab:main_results}. \emph{First Utterance} refers to a naive model where the first utterance after the question is considered as the answer.

As can be seen in \tabref{tab:main_results}, using the FHL objective, significantly improves performance when no answer is available (\eg 75.5 \vs 69.6 in EM and 80.9 \vs 76.0 in F1 with DeBERTa). When the answer is available, results remain mostly unchanged. The results observed demonstrate the effectiveness of FHL in improving performance of any model when the data contains a significant proportion of unanswerable questions as is the case with MeeQA. 
Note that overall results are much worse compared to other standard QA datasets like SQuAD which leaves opportunities for improvement.
Importantly, humans are still much better at detecting unanswerable questions, as our results suggest.

\noindent\textbf{Ablation Analysis}
For this experiment we consider three variants of FHL, obtained by removing one of the terms defined in \eqref{eq:hierarchical_loss} (each term is preceded by a scalar $\alpha$, $\beta$ or $\gamma$ in \eqref{eq:hierarchical_loss}). Result of the ablation analysis are reported in \tabref{tab:ablation_results}. Importantly, removal of the loss term responsible for \emph{answerability} prediction, \ie, ``No-HA'', drops performance by over 25\% in F1 and EM. Hence, incorporating it into FHL is crucial.

\vspace{-0.2cm}
\section{Conclusions}
\label{conclusions}
\vspace{-0.2cm}
In this paper, we introduce MeeQA, a large-scale dataset for extracting answers to natural questions in meeting transcripts. Unlike existing RC and QA datasets, MeeQA contains natural questions from real meetings, along with their answers, if they exist. Furthermore, to improve existing model performances on unanswerable questions, we propose a novel loss function designed to directly handle these type of questions. Our experiments show that using FHL yield models that outperform others, especially on unanswerable questions. We hope to explore other use cases for this type of loss in the future. However, human performance on these questions is still better than our model. We hope this work will encourage other researchers to explore this task, which is critical for extracting meaningful information from meetings.

\section*{Ethics Consideration}
We propose a novel meeting transcripts question answering task, which is accompanied by a dataset MeeQA. Both the transcripts of the meetings and the annotators recruited touch on the intellectual property rights and privacy rights of the original authors. As part of the dataset construction process, we ensure that the intellectual property and privacy rights of the original authors of the meetings are protected. All the meeting transcriptions we collected are publicly available. Additionally, the annotation process complies with the intellectual property and privacy rights of the recruited annotators.


\bibliography{anthology,custom}
\bibliographystyle{acl_natbib}

\clearpage
\appendix
\section*{Supplementary Material}
\label{sec:appendix}
\begin{figure*}[t]
\includegraphics[width=1\linewidth]{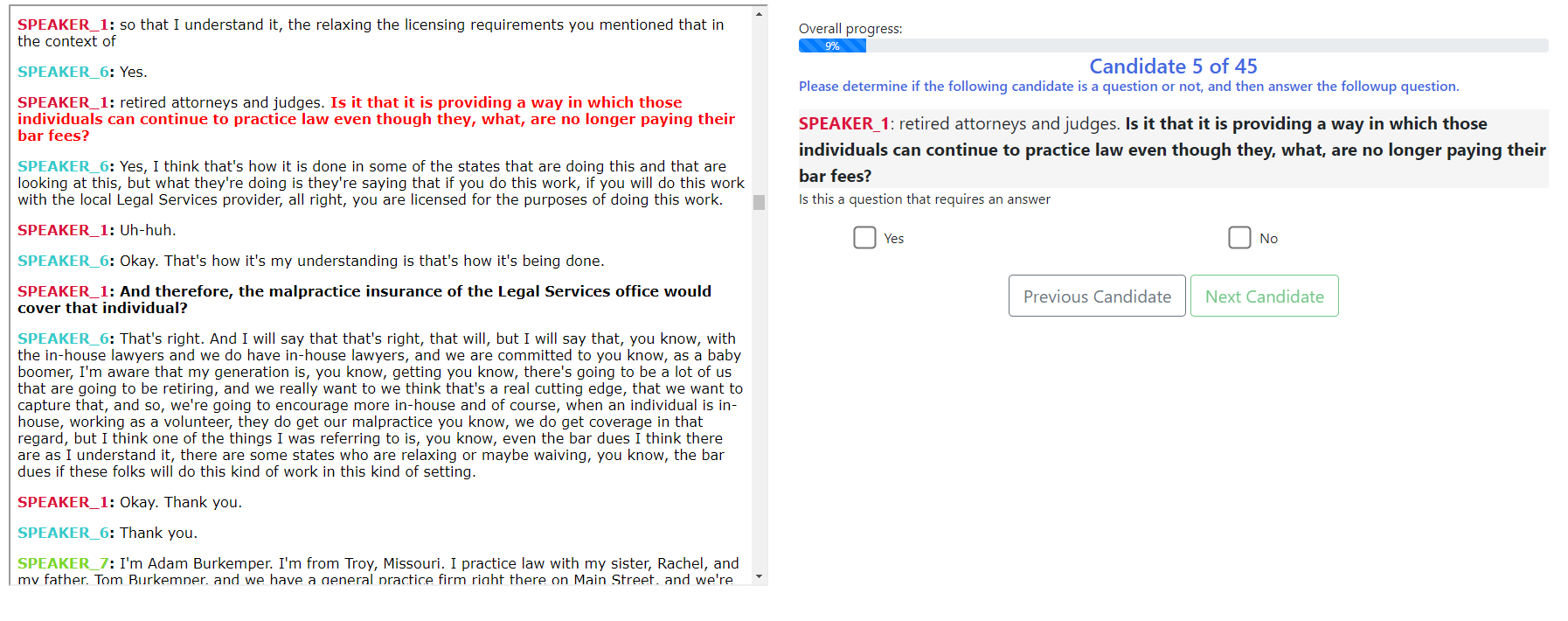}
\caption{The judges first question}
\label{fig:first_screen}

\includegraphics[width=1\linewidth]{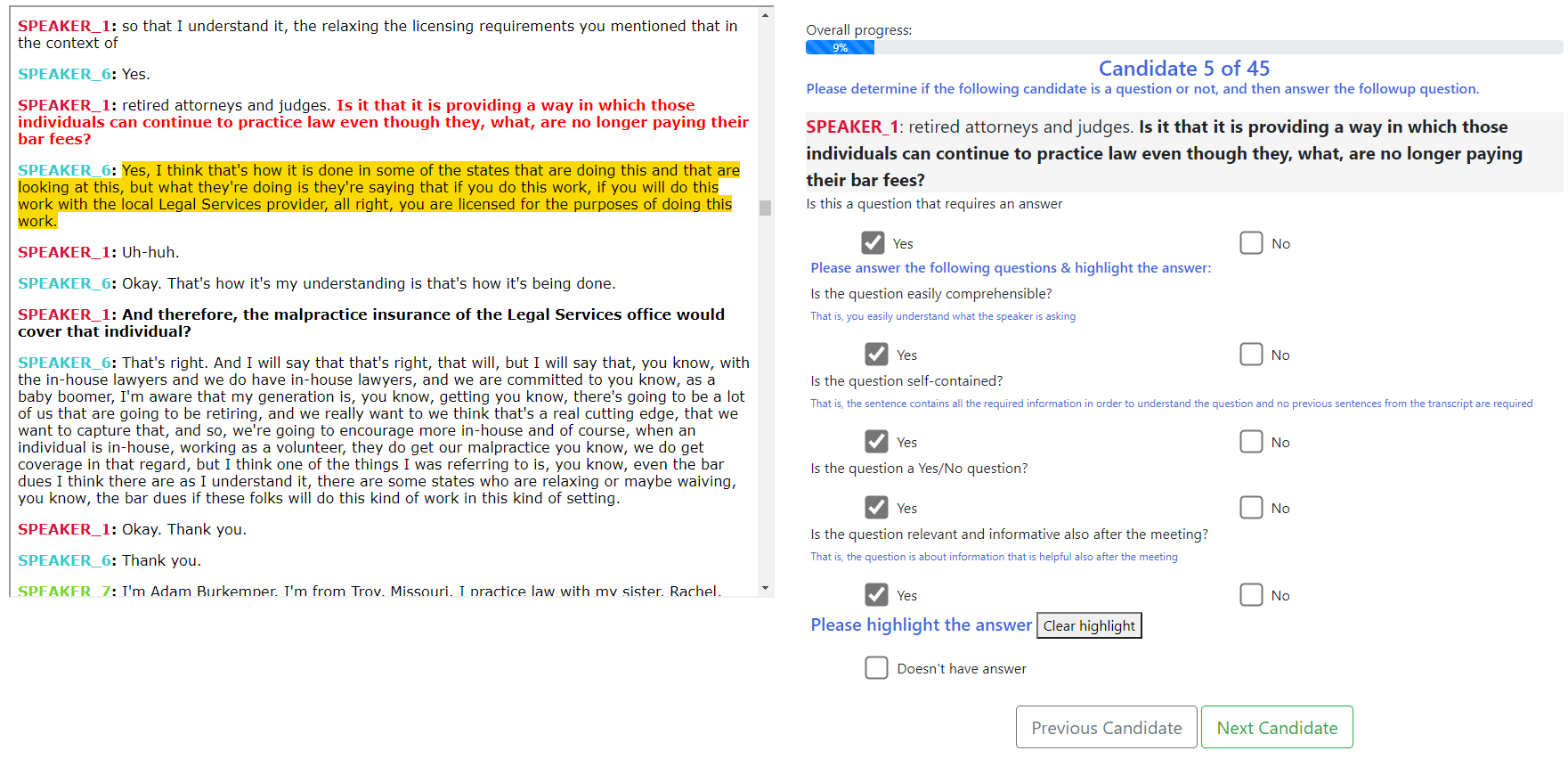}
\caption{Questions for the judges in the event of an affirmative answer to the first question}
\label{fig:second_screen}
\end{figure*}

This supplementary material provides the
judges annotation interfaces in
Figure \ref{fig:first_screen} and Figure \ref{fig:second_screen}. We also give details regarding the data pre-processing in Appendix \ref{app:data_pre_processing} and the data representation in Appendix \ref{app:data_representation}.

\section{Data Pre-Processing}
\label{app:data_pre_processing}

As the text in MeeQA is a spoken language converted to text, we pre-process the data before inserting it into the models.  Our data pre-processing includes several steps. We clean the data from comments that aim to indicate something happened during the meeting. Square or triangular brackets mark these comments, for example <laugh> and [No response.]. We also remove comments that were added to the transcript, like``secretary's note : this statement was not found in committee records.''. Moreover, we remove filler words like ``uh'' or ``hmm'' and words repetition like ``Well it is it is more...''. Finally, we remove special signs like @ and \_, brackets, and one-character words.

We then complete words that were partially highlighted, and we exclude from the dataset question-answer pairs in which the answer is in the utterances before the question.
In the case of a multi-span answer, we merge the two spans if only one word separated them, and one of the spans speakers told it. 

\section{Data Representation}
\begin{figure}[th!]
\footnotesize
\centering
\begin{tabular}{p{\columnwidth}}
\toprule
\texttt{$u_{q-2}$}: CHAIRMAN McKAY: And so this is a discussion that will continue. So unless I hear a screaming objection, I ask that we move onto the next item on the agenda. \\
\texttt{$u_{q-1}$}: MR. POLGAR: Okay. Thank you.\\
\texttt{$u_q$}: CHAIRMAN McKAY: Thank you very much. The next item on the agenda is considering and acting on the change of address notification to diversify investment advisors. Who is going to speak on that?\\
\texttt{$u_{q+1}$}: CHAIRMAN McKAY: Mr. Jeffress, I suspect this will be a short.\\
\midrule
\textbf{\textsc{Original}}: SPEAKER\_{0}: Thank you very much. The next item on the agenda is considering and acting on the change of address notification to diversify investment advisors. Who is going to speak on that? \texttt{[SEP]} SPEAKER\_{0}: Mr. Jeffress, I suspect this will be a short.\\
\vspace{0em}
\textbf{\textsc{Switch Speakers+Original}}:SPEAKER\_{0}: Thank you very much. The next item on the agenda is considering and acting on the change of address notification to diversify investment advisors. Who is going to speak on that? \texttt{[SEP]} Mr. Jeffress, I suspect this will be a short. \\
\vspace{0em}
\textbf{\textsc{Switch speakers+One previous utterance}}:  SPEAKER\_{1}: Okay. Thank you. SPEAKER\_{0}: Thank you very much. The next item on the agenda is considering and acting on the change of address notification to diversify investment advisors. Who is going to speak on that? \texttt{[SEP]} Mr. Jeffress, I suspect this will be a short.\\
\vspace{0em}
\textbf{\textsc{Switch speakers+Two previous utterance}}: SPEAKER\_{0}: And so this is a discussion that will continue. So unless I hear a screaming objection, I ask that we move onto the next item on the agenda. SPEAKER\_{1}: Okay. Thank you. SPEAKER\_{0}: Thank you very much. The next item on the agenda is considering and acting on the change of address notification to diversify investment advisors. Who is going to speak on that? \texttt{[SEP]} Mr. Jeffress, I suspect this will be a short.\\
\bottomrule
\end{tabular}
\caption{A partial meeting transcript and different speaker and question representations of the question.}
\label{fig:example_question_representation}
\end{figure}
\label{app:data_representation}
We consider three different options for representing a transition between two consecutive utterances and representing the question used as an input for all the models. \figref{fig:example_question_representation} presents an example of the different representations for a sequence.

\begin{itemize}
    \item \textbf{Original:} This option serves as a baseline where the data and the questions are represented in their original format. A special token separates every two utterances, ``\&'', followed by the speaker token (as defined in  \cref{subsec:model}), and we pass the model just the question, without any previous utterances, i.e., $k=0$.
    \item \textbf{Switch Speakers:} In this representation, we use the token representing a new utterance and the speaker token only when the speaker of the current utterance is not the same as the one of the previous one. Regarding the question representation, we consider two options:
    \begin{itemize}
        \item[$\ast$] \textbf{Original:} We pass the model just the question without previous utterances, i.e., $k=0$.
        \item[$\ast$] \textbf{Previous Utterances:} We pass the model the question as a concatenation of one or two utterance before it and the question itself, i.e., $k \in (1, 2)$.
    \end{itemize}
\end{itemize}

\tabref{tab:data_representation} shows the results of our models for each data representation option.

\begin{table*}[]
\centering
\begin{tabular}{lllllllll}
\toprule
\multicolumn{1}{c|}{\multirow{2}{*}{\textbf{Model}}}&\multicolumn{1}{c}{\multirow{2}{*}{\textbf{\begin{tabular}[c]{@{}c|@{}}Speaker\\Representation\end{tabular}}}}&\multicolumn{1}{c|}{\multirow{2}{*}{\textbf{\begin{tabular}[c]{@{}c@{}}Question\\Representation\end{tabular}}}}&\multicolumn{2}{c}{\textbf{AllData}}&\multicolumn{2}{c}{\textbf{HasAns}}&\multicolumn{2}{c}{\textbf{NoAns}}\\\cline{4-9}
\multicolumn{1}{c|}{}&\multicolumn{1}{c}{}&\multicolumn{1}{c|}{}&\textbf{EM}&\textbf{F1}&\textbf{EM}&\textbf{F1}&\textbf{EM}&\textbf{F1}\\
\midrule
\multicolumn{9}{c}{\textbf{Dev}}\\\midrule
\multicolumn{1}{l|}{\multirow{3}{*}{\textbf{BERT+FHL}}}&\multicolumn{1}{l|}{\textbf{Original}}&\multicolumn{1}{l|}{\textbf{Original}}&65.3&74.0&62.5&72.5&82.1&83.0\\
\multicolumn{1}{l|}{}&\multicolumn{1}{l|}{\textbf{Switch}}&\multicolumn{1}{l|}{\textbf{Original}}&63.3&75.6&60.2&74.2&82.4&84.6\\
\multicolumn{1}{l|}{}&\multicolumn{1}{l|}{\textbf{Switch}}&\multicolumn{1}{l|}{\textbf{$k=1$}}&62.9&75.7&60.2&74.6&79.6&82.5\\
\multicolumn{1}{l|}{}&\multicolumn{1}{l|}{\textbf{Switch}}&\multicolumn{1}{l|}{\textbf{$k=2$}}&62.6&75.4&59.6&74.0&81.1&83.5\\\hline
\multicolumn{1}{l|}{\multirow{3}{*}{\textbf{DeBERTa+FHL}}}&\multicolumn{1}{l|}{\textbf{Original}}&\multicolumn{1}{l|}{\textbf{Original}}&16.1&69.5&5.2&66.7&83.4&87.0\\
\multicolumn{1}{l|}{}&\multicolumn{1}{l|}{\textbf{Switch}}&\multicolumn{1}{l|}{\textbf{Original}}&54.8&76.1&50.4&74.5&82.3&85.9\\
\multicolumn{1}{l|}{}&\multicolumn{1}{l|}{\textbf{Switch}}&\multicolumn{1}{l|}{\textbf{$k=1$}}&53.9&76.0&50.0&74.8&77.9&83.2\\
\multicolumn{1}{l|}{}&\multicolumn{1}{l|}{\textbf{Switch}}&\multicolumn{1}{l|}{\textbf{$k=2$}}&53.3&75.5&49.2&74.4&78.1&82.1\\\hline
\multicolumn{1}{l|}{\multirow{3}{*}{\textbf{Albert+FHL}}}&\multicolumn{1}{l|}{\textbf{Original}}&\multicolumn{1}{l|}{\textbf{Original}}&65.7&74.1&62.4&72.0&85.9&86.5\\
\multicolumn{1}{l|}{}&\multicolumn{1}{l|}{\textbf{Switch}}&\multicolumn{1}{l|}{\textbf{Original}}&63.4&76.1&59.8&74.3&85.4&87.5\\
\multicolumn{1}{l|}{}&\multicolumn{1}{l|}{\textbf{Switch}}&\multicolumn{1}{l|}{\textbf{$k=1$}}&64.0&76.4&60.3&74.5&86.5&88.0\\
\multicolumn{1}{l|}{}&\multicolumn{1}{l|}{\textbf{Switch}}&\multicolumn{1}{l|}{\textbf{$k=2$}}&63.1&75.7&59.8&74.0&83.4&86.0\\\midrule
\multicolumn{9}{c}{\textbf{Test}}\\\midrule
\multicolumn{1}{l|}{\multirow{3}{*}{\textbf{BERT+FHL}}}&\multicolumn{1}{l|}{\textbf{Original}}&\multicolumn{1}{l|}{\textbf{Original}}&65.3&73.8&63.3&72.9&78.4&79.2\\
\multicolumn{1}{l|}{}&\multicolumn{1}{l|}{\textbf{Switch}}&\multicolumn{1}{l|}{\textbf{Original}}&64.0&75.7&61.5&74.6&79.2&82.1\\
\multicolumn{1}{l|}{}&\multicolumn{1}{l|}{\textbf{Switch}}&\multicolumn{1}{l|}{\textbf{$k=1$}}&62.3&74.7&60.8&74.4&72.2&76.9\\
\multicolumn{1}{l|}{}&\multicolumn{1}{l|}{\textbf{Switch}}&\multicolumn{1}{l|}{\textbf{$k=2$}}&62.9&74.8&60.9&74.0&75.6&79.6\\\hline
\multicolumn{1}{l|}{\multirow{3}{*}{\textbf{DeBERTa+FHL}}}&\multicolumn{1}{l|}{\textbf{Original}}&\multicolumn{1}{l|}{\textbf{Original}}&16.1&69.5&5.6&67.0&82.0&85.1\\
\multicolumn{1}{l|}{}&\multicolumn{1}{l|}{\textbf{Switch}}&\multicolumn{1}{l|}{\textbf{Original}}&56.2&75.7&52.2&74.2&81.1&85.3\\
\multicolumn{1}{l|}{}&\multicolumn{1}{l|}{\textbf{Switch}}&\multicolumn{1}{l|}{\textbf{$k=1$}}&55.2&75.2&52.0&74.3&75.5&80.9\\
\multicolumn{1}{l|}{}&\multicolumn{1}{l|}{\textbf{Switch}}&\multicolumn{1}{l|}{\textbf{$k=2$}}&54.8&75.8&51.5&75.0&75.3&80.7\\\hline
\multicolumn{1}{l|}{\multirow{3}{*}{\textbf{Albert+FHL}}}&\multicolumn{1}{l|}{\textbf{Original}}&\multicolumn{1}{l|}{\textbf{Original}}&65.3&73.6&63.0&72.4&79.9&80.8\\
\multicolumn{1}{l|}{}&\multicolumn{1}{l|}{\textbf{Switch}}&\multicolumn{1}{l|}{\textbf{Original}}&63.3&75.3&60.5&73.9&81.0&84.3\\
\multicolumn{1}{l|}{}&\multicolumn{1}{l|}{\textbf{Switch}}&\multicolumn{1}{l|}{\textbf{$k=1$}}&63.2&75.5&60.4&74.1&80.7&84.4\\
\multicolumn{1}{l|}{}&\multicolumn{1}{l|}{\textbf{Switch}}&\multicolumn{1}{l|}{\textbf{$k=2$}}&63.5&75.0&60.9&73.8&79.4&83.0\\
\bottomrule
\end{tabular}
\caption{Data representation results. Performance of models with FHL trained over different data representation on development and test data.}
\label{tab:data_representation}
\end{table*}

\end{document}